\documentclass[letterpaper, 10 pt, conference]{ieeeconf}  
\IEEEoverridecommandlockouts 
\newif\ifblindreview

\usepackage{cite}
\usepackage{amsmath,amssymb,amsfonts}

\usepackage{graphicx}
\usepackage{textcomp}
\usepackage{booktabs}
\usepackage{xcolor}
\usepackage{soul}
\usepackage{caption}
\usepackage{subcaption}

\usepackage{dblfloatfix}
\usepackage{array}
\usepackage{url}
\usepackage{etoolbox}
\makeatletter
\patchcmd{\@maketitle}
  {\vskip1.0em\par}
  {\vskip0.1em\par}
  {\typeout{Title spacing patch successful}}
  {\typeout{WARNING: Title spacing patch failed}}
\makeatother
\def\BibTeX{{\rm B\kern-.05em{\sc i\kern-.025em b}\kern-.08em
    T\kern-.1667em\lower.7ex\hbox{E}\kern-.125emX}}

\newcount\Comments %
\usepackage{color}
\newcommand{\kibitz}[2]{\ifnum\Comments=1\textcolor{#1}{#2}\fi}

\title{\LARGE \bf OUCH: Objective, Uniform, Controlled Haptic Stimulation\\ for Automated Neurological Assessment}

\ifblindreview
  \author{Author names and affiliations withheld for double anonymous review
  }
\else
  \author{
    Sherry Choi$^{1,2, \dagger}$,
    Sitao Sunny Li$^{1, \dagger}$,
    Joseph Romero$^{3}$,\\
    Lillian Chin$^{3}$,
    Ashwin G. Ramayya$^{4}$,
    Allison M. Okamura$^{1}$,
    and Ava Chen$^{1}$
  \thanks{This work was supported in part by the Stanford Coulter Translational Research Program, the Stanford Wu Tsai Institute Neuroscience:Translate Program, and the Rice University Owl Edge Summer Experience Fund.}
    \thanks{$\dagger$ These authors contributed equally.}
    \thanks{$^{1}$Department of Mechanical Engineering, Stanford University, Stanford, CA, USA. \texttt{\{sunnystl, aokamura, avachen\}@stanford.edu}} %
    \thanks{$^{2}$Department of Mechanical Engineering, Rice University, Houston, TX, USA. \texttt{sc233@rice.edu}}%
    \thanks{$^{3}$Department of Electrical and Computer Engineering, University of Texas at Austin, TX, USA.}%
    \thanks{$^{4}$Department of Neurosurgery, Stanford University, Stanford, CA, USA.}%
}
\fi

\begin{document}

\maketitle
\thispagestyle{empty}
\pagestyle{empty}
\vspace{-1.0em}

\begin{abstract}
Motor responses evoked by noxious stimulation are used to assess neurological function in hospitalized patients with disorders of consciousness. In standard bedside assessments, these stimuli are applied manually, resulting in substantial variation in force, duration, frequency, and location. Here, we present OUCH (Objective, Uniform, Controlled Haptic stimulation), a wrist-worn device that uses pneumatic actuation to deliver noxious mechanical stimuli. OUCH combines stacked pneumatic pouch actuators, a rigid indenter, and a compliant spacer designed to buckle at a set load, so that the indenter contacts the skin only during actuation. We characterize the device using blocked force tests, clinician evaluation, finite element analysis, and a deformable pressure testbed, and benchmark it against a commercially available handheld pressure algometer. At 200~kPa pouch inflation pressure, our indenter concentrated its load into an estimated contact pressure of 18,500~kPa, over 25 times that of the algometer tip, while requiring less total force. On the deformable testbed, it localized deformation near the surface, while deformation at depth remained comparable to the algometer. As a result, in preliminary perceptual testing, our indenter elicited discomfort at approximately 89\% lower applied force than the algometer, demonstrating the feasibility of evoking discomfort with a compact, wearable soft actuator. These results show that combining pneumatic actuation with designed contact geometry enables controlled, repeatable noxious stimulation at the wrist, providing a foundation for objective neurological assessment.
\end{abstract}

\section{Introduction}
Motor response provides one of the few observable windows into neurological function in patients with disorders of consciousness, who are often unable to communicate or follow commands~\cite{bodien2026}. Current standard bedside assessments for this population, such as the Glasgow Coma Scale (GCS) and the Coma Recovery Scale-Revised (CRS-R), evaluate motor response by a clinician manually delivering a noxious localized pinch or pressure stimulus, then subjectively categorizing the evoked movement~\cite{Schnakers2024, teasdale2014}. More reflexive responses indicate lower neurological function, and more purposeful movements indicate higher function~\cite{nallaluthan2023, lower2010}. However, the high variance in force, duration, frequency, and location of manually applied stimuli across examiners and sessions~\cite{cook2019, teasdale2014}, as well as the subjective interpretation of the evoked response, present a challenge for monitoring changes in neurological state~\cite{Schnakers2024, edlow2021, norman2010}.

\begin{figure}[!t]
    \centering
    \vspace{2mm}
    \includegraphics[width=\linewidth]{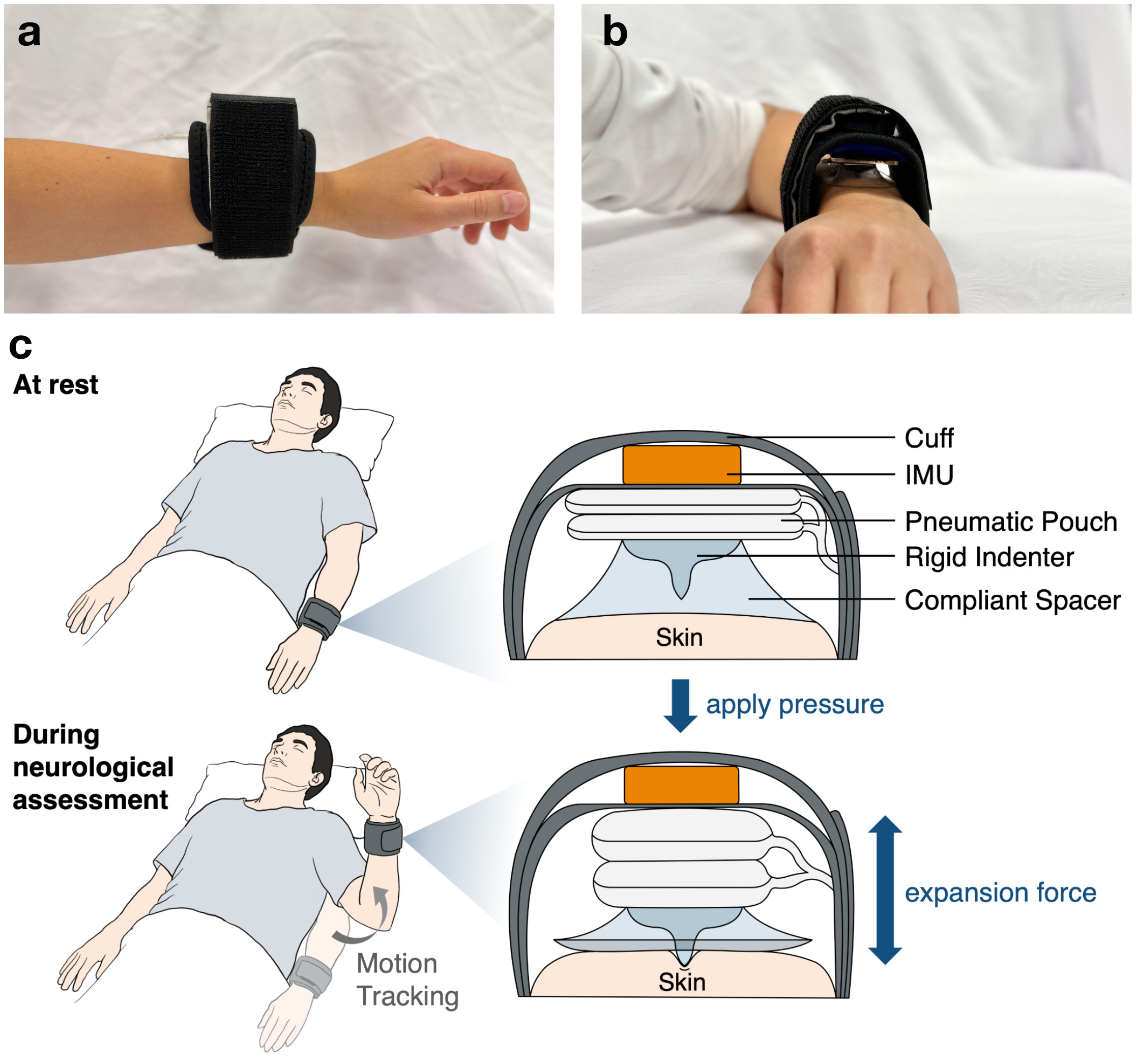}
    \captionsetup{font=footnotesize}
    \caption{\textbf{OUCH wearable device and operating principle.}
    (a) Side view and
    (b) front view of OUCH on the wrist.
    (c) Schematic of the device at rest and during neurological assessment, where pneumatic pouch actuator expansion deforms the compliant spacer and drives the indenter into contact with the skin while the IMU tracks motion.}
    \label{fig:ouch_actuator}
    \vspace{-7mm}
\end{figure}

Several approaches have been developed to deliver more controlled mechanical stimuli. Handheld pressure algometers, such as the AlgoMed (Medoc Ltd.), display the applied force and are commonly used to measure pain thresholds. However, these devices require an examiner to manually position the device and apply force against the body~\cite{algomed}. Many compact pneumatic wearables have been developed to apply localized pressures to the skin~\cite{dupasquier2024}, although these are typically designed for innocuous touch and generate sensations well below pain thresholds. Polianskis et al. measured pressure-pain responses with a computer-controlled pneumatic tourniquet cuff, but their device produced a deep squeeze stimulus rather than the localized pain used to evoke movement in clinical neurological assessment~\cite{polianskis2001}. Other human studies using controlled indentation and impact experiments have shown that small contacts can evoke pain at low forces~\cite{rolke2006}, and that the onset of perceived pain occurs at lower contact pressures than blunt tissue injury~\cite{behrens2022, behrens2023} or skin penetration~\cite{shergold2005}, suggesting that noxious stimulation need not be injurious. This presents an opportunity to deliver a safe, localized stimulus with a wearable device in a controlled and uniform manner, eliciting a response to the noxious stimulus without causing tissue injury.

In this work, we present OUCH (Objective, Uniform, Controlled Haptic stimulation), a wrist-worn pneumatic device for controlled noxious stimulation (Fig.~\ref{fig:ouch_actuator}). OUCH combines two widely studied mechanisms in soft robotics---textile pouch actuators and tunable-buckling silicone structures---to adapt pneumatic wearables for noxious stimulation. First, we introduce the design of each component of the device with the goal of delivering localized pain to the body only when the device is inflated. Second, we experimentally quantify how inflation pressure relates to force, contact pressure, and deformation produced by our indenter design on rigid and deformable surfaces, and to perceived discomfort thresholds. Finally, we compare the mechanical response of our indenter geometry against that of the indenter geometry of a clinical handheld algometer. Together, these experiments characterize the force and contact pressure that a soft wearable can generate to evoke a pain response at the wrist, revealing that localizing the load is more effective than increasing force.

\section{Device Design}
We designed OUCH as a soft, snug wrist-mounted device that delivers controlled indentation force over a small contact area using relatively low pneumatic inflation pressure. To achieve this, the device (Fig.~\ref{fig:ouch_actuator}) combines a pneumatic textile pouch actuator, a rigid indenter, a compliant spacer, and a fabric wrist cuff that secures the device to the user. We selected a pneumatic pouch actuator because its thin fabric construction enables integration into a compact, soft wrist-worn device, while its inflation pressure provides a simple means of controlling force output. As the pouch inflates, it drives the rigid indenter against the skin to localize the mechanical pain stimulus at a small region on the dorsal side of the wrist. To avoid unintended discomfort and ensure that the rigid indenter only contacts the skin during actuation, we designed a compliant silicone spacer to suspend the indenter away from the wrist while the device is inactive and to buckle during actuation to allow the indenter to contact the skin. The pouch is inserted into a dedicated stretchable pocket inside the wrist cuff, while the indenter and compliant spacer are positioned beneath the pouch and secured to the pocket using Velcro. The cuff also positions the pouch beneath an integrated IMU wrist-motion sensor, which provides a rigid surface for the pouch to react against during inflation.

\subsection{Pneumatic Pouch Actuator Design}
\label{sec:pouch}
We fabricated the pouch actuator from TPU-coated 70-denier ripstop nylon (Quest Outfitters \#1056) using ultrasonic welding (Vetron 5064) to form airtight seams. Before welding, we poked a small hole in the flat pouch surface and inserted a 2~mm polypropylene barb fitting (Gikfun \#EK1260) for air connection. We sealed the fabric-barb interface using silicone sealant (Seam Grip SIL, GEAR AID) and an approximately 1~mm thick silicone ring clamp. After sealing, we cured the assembly for 24~h before testing. We fabricated and evaluated several pouch geometries to investigate how pouch construction affected force generation and reliability.

\begin{figure}[t]
    \centering
    \includegraphics[width=0.95\linewidth]{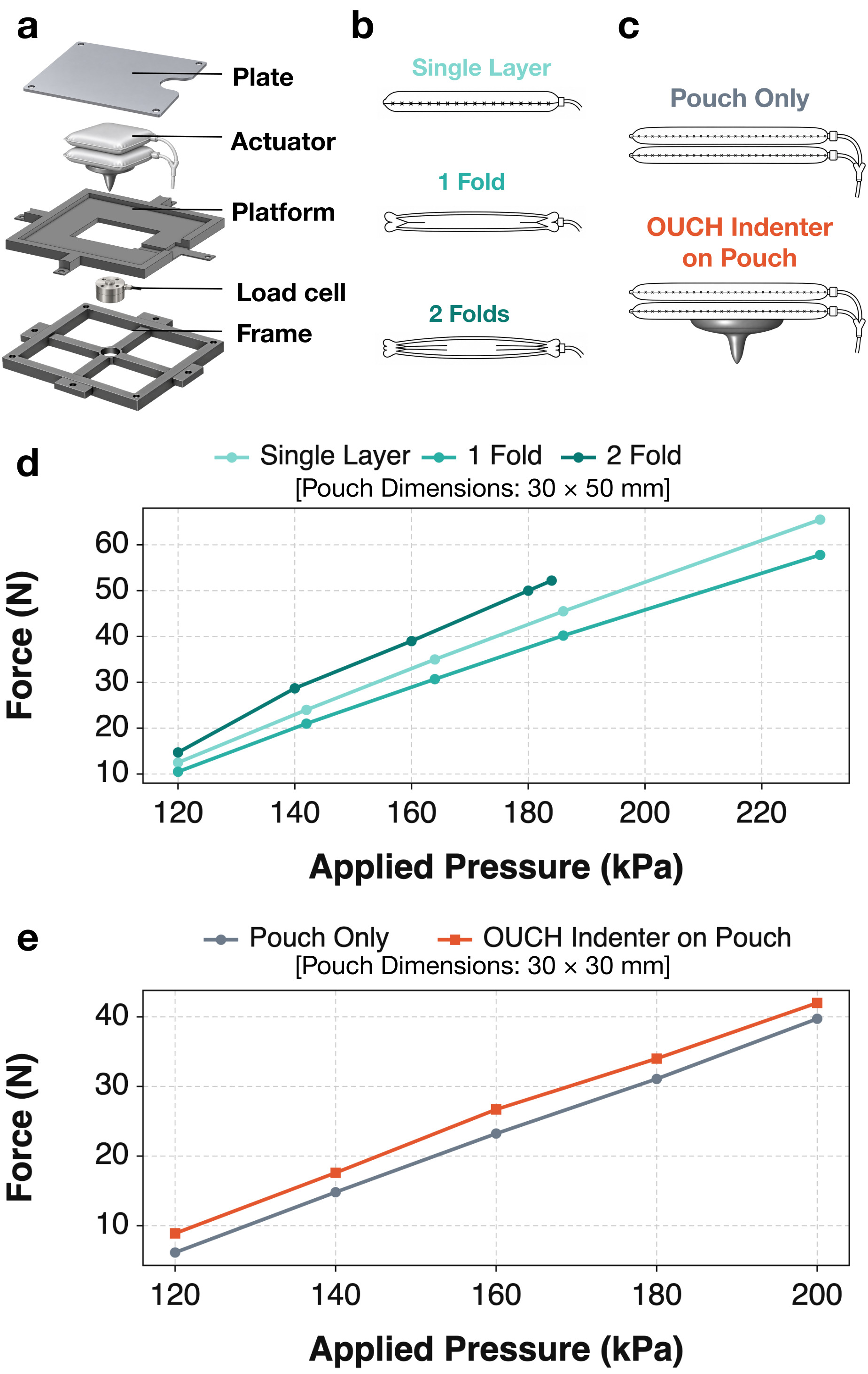}
    \captionsetup{font=footnotesize}
    \caption{\textbf{Pouch design and blocked force characterization.}
    (a) Exploded view of the blocked force test setup.
    (b) Pneumatic pouch actuator design iterations with increasing numbers of folds.
   (c) Finalized stacked 2$\times$1 single-layer pouch configuration shown without the OUCH indenter (pouch only; top) and with the OUCH indenter attached beneath the pouch stack (bottom).
    (d) Blocked force comparison of the pouch designs shown in (b).
    (e) Blocked force comparison of the finalized pouch configuration with and without the OUCH indenter}
    \label{fig:pouch_indenter}
    \vspace{-8mm}
\end{figure}

To quantify the force output of each pouch configuration with inflation pressure, we conducted a blocked force test using a designed test rig~\cite{coram2025sealing}. As shown in Fig.~\ref{fig:pouch_indenter}(a), the actuator is placed on a 3D-printed platform with a designated hole for the actuator to expand through. When inflated, the actuator presses against a load cell (ATI Nano17 6-axis Force/Torque Sensor) beneath it. Above the actuator, a rigid acrylic plate prevents the actuator from expanding away from the load cell. We tested pressures from 120~kPa, which exceeded ambient pressure and initiated pouch expansion, to 230~kPa, the maximum pressure reached before pouch leakage. The load cell recorded the blocked force at each pressure for comparison between pouch designs.

We investigated whether adding folds could increase pouch force output by comparing single-layer, 1-fold, and 2-folds designs. Fig.~\ref{fig:pouch_indenter}(b) shows the three designs, and Fig.~\ref{fig:pouch_indenter}(d) compares their blocked force performance. To keep the actuator compact, we set the planar dimensions of the initial designs to 30~mm $\times$ 50~mm, the smallest geometry we could reliably fabricate given the complexity of the folded configurations. The single-layer configuration consisted of a perimeter-sealed pouch without additional folds, whereas the 1-fold and 2-folds configurations incorporated one or two 15~mm $\times$ 50~mm inward folds, respectively. We expected the added material to provide greater internal volume and out-of-plane expansion, thereby increasing force output at a given inflation pressure. Blocked force increased with applied pressure for all three configurations. The single-layer and 1-fold pouches were tested at five evenly spaced pressure levels between 120 and 230~kPa, while leakage limited the 2-folds pouch to pressures between 120 and 184~kPa. At 180~kPa, the single-layer, 1-fold, and 2-folds pouches generated 45.5, 42.6, and 52.3~N, respectively. The 2-folds result supported our expectation, but the 1-fold pouch unexpectedly generated less force than the single-layer configuration. Thus, force output did not increase monotonically with the number of folds. Asymmetric unfolding, differences in weld geometry or sealing quality, and air leakage may have reduced the effective transmission of pressure into normal force.

To increase actuation capacity while reducing the fabrication complexity and leakage risk introduced by additional folds, we stacked multiple single-layer pouches and supplied each with a combined air source. We assembled the pouches within a stretchable fabric casing that maintained their alignment while allowing them to expand during actuation. Because the individual pouches no longer required folds, the stacked architecture simplified fabrication and allowed us to reduce their planar dimensions. We standardized the final design to use two 30~mm $\times$ 30~mm pouches, as shown in Fig.~\ref{fig:pouch_indenter}(c). Additional pouch layers can be stacked if greater force output is required. At an applied pressure of 200~kPa, the final two-layer configuration generated a blocked force of 39.7~N.

\subsection{Rigid Indenter Design}

We coupled a rigid indenter, dubbed the OUCH indenter, to the pneumatic pouch actuator to concentrate the actuator force over a small contact area and amplify contact pressure at the wrist. We defined the indenter geometry as a revolved spline profile, forming an axisymmetric, narrow, rounded tip with a contact area of 2.27~mm$^2$. The tip transitions smoothly into a wider base with a diameter of 17~mm, and the overall indenter height is 8~mm. We 3D printed the indenter in PLA using fused-filament fabrication (Bambu Lab X1C). To determine whether adding the indenter amplified the transmitted force, we conducted blocked force tests with and without it (Fig.~\ref{fig:pouch_indenter}(e)). Across the tested pressure range of 120 to 200~kPa, both configurations exhibited increasing blocked force with increasing applied pressure, and the pouch with the OUCH indenter generated a consistently higher blocked force, by approximately 2.5~N. At 200~kPa, the configuration with the OUCH indenter generated 42.0~N. These results indicate that the OUCH indenter maintained comparable blocked force output while producing a more localized pressure distribution at the contact interface.

We collected preliminary clinician feedback to evaluate the perceived stimulus produced by the pneumatic pouch actuator alone and by the pouch with the OUCH indenter attached. During the evaluation, clinicians wore the fully assembled device on the wrist while we inflated the pouch to 200~kPa. We asked them to describe the overall sensation produced by each configuration and to assess whether the stimulus was sufficiently noxious for use in neurological assessment. Table~\ref{tab:clinician_feedback} presents representative verbatim quotes from their feedback. Overall, clinicians indicated that the pouch with the OUCH indenter produced a stimulus comparable in intensity to those used in current bedside neurological assessments, making it more suitable for eliciting a response to noxious stimulation than the pouch alone.

\begin{table}[t]
    \vspace{1mm}
    \centering
   \caption{\textbf{Representative clinician feedback comparing the pneumatic pouch actuator alone and the pouch with the OUCH indenter.}}
    \label{tab:clinician_feedback}

    \begin{tabular}{
        >{\raggedright\arraybackslash}p{0.20\linewidth}
        >{\raggedright\arraybackslash}p{0.45\linewidth}
        >{\raggedright\arraybackslash}p{0.22\linewidth}
    }
        \hline
        \textbf{Condition} &
        \textbf{Representative Quotes} &
        \textbf{Perceived Sensation} \\
        \hline

        Pouch Only &
        ``It doesn't hurt me." \newline
        ``Feels like I've overtightened the watch strap." &
        Broad pressure sensation without pain. \\
        \hline

        Pouch with OUCH Indenter &
        ``Oh now it's starting to hurt." \newline
        ``This is perfect." \newline
        ``This is good, I like it." &
        Noticeable discomfort at an appropriate intensity. \\
        \hline
    \end{tabular}
    \vspace{-7mm}
\end{table}

\subsection{Compliant Spacer Design}
We placed a compliant spacer between the OUCH indenter and the skin to hold the indenter away from the skin when the actuator is not applying a stimulus. The spacer resembles a suction cup, and includes a thick backing pad distributing pouch inflation pressure to the indenter base and thin walls that widen from a narrower neck to a broader circular rim. Once the load exceeds the buckling threshold, the walls widen and curl upward, allowing the indenter to contact the skin and deliver the stimulus.

\begin{figure}[t]
    \centering
    \includegraphics[width=0.9\linewidth]{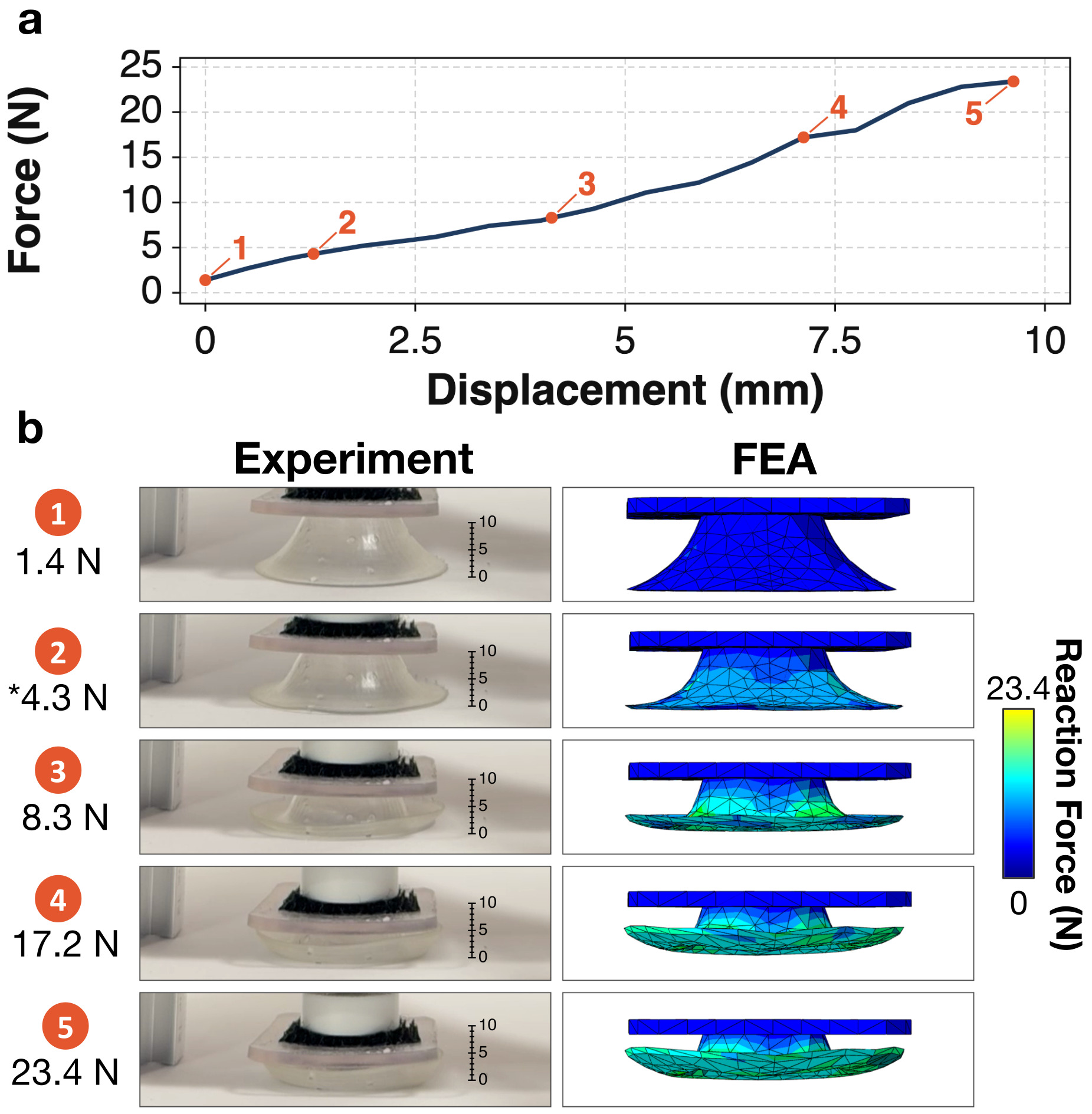}
    \captionsetup{font=footnotesize}
    \caption{\textbf{Compliant spacer design and characterization.}
    (a) Measured force-displacement response during compression of the spacer, with numbered points corresponding to selected deformation states.
    (b) Comparison of spacer deformation during manual compression testing and finite element analysis (FEA) at the corresponding applied forces. Asterisk (second) state is the buckling point.}
    \label{fig:spacer_fea}
    \vspace{-7mm}
\end{figure}

We 3D printed the spacer from Flexible 80A resin (Formlabs), a TPU-like elastomer with a Shore hardness of 80A, and tuned the wall thickness and curve geometry to achieve the desired buckling onset behavior. To characterize the spacer’s mechanical behavior, we performed finite element analysis (FEA, Abaqus/Standard Learning Edition 2025) and compared the results with qualitative observations from manual compression. Fig.~\ref{fig:spacer_fea}(b) shows the FEA model, in which the spacer was compressed against an analytical rigid surface using a 10~mm downward displacement. Contact was modeled using finite sliding with a friction coefficient of 0.8, and the spacer was assigned a Young’s modulus of 3.5~MPa and a Poisson’s ratio of 0.48. We defined the onset of buckling as the point at which the simulated reaction force reached a local peak. The force peaked at 4.3~N at 2.67~mm of downward displacement, marking the transition into the buckled configuration.

We then compared the simulated deformation states with those observed during a manual compression test. For this test, we used a handheld force sensor to gradually compress the spacer while it rested on a rigid tabletop. The measured force-displacement response in Fig.~\ref{fig:spacer_fea}(a) showed a similar buckling transition at 4.3~N and 2.5~mm of compression. The agreement in both the buckling force and the deformation shape suggests that the FEA model captured the primary mechanical behavior of the fabricated spacer.

\section{Experimental Characterization}

With the OUCH design finalized in Section~II, we next characterized the fully assembled OUCH device and benchmarked it against the AlgoMed pressure algometer, a clinically used reference for quantitative pressure-pain assessment. The AlgoMed tip is a flat cylindrical contactor with a diameter of 10~mm (contact area 78.5~mm$^2$) and a height of 5~mm. For all comparisons, we used a PLA replica of the AlgoMed tip geometry mounted on the same stacked-pouch actuator in place of the OUCH indenter. Because the OUCH indenter is intended for soft skin contact, we first developed a deformable pressure testbed that serves as a tissue proxy, along with a finite element model of OUCH acting on a deformable surface. Using the testbed, we compared how the two indenters deform the surface at different depths, and used an FEA model to examine contact interactions that cannot be directly observed experimentally. We also measured force output under rigid blocked force conditions and estimated the corresponding contact pressures. Finally, we examined how these mechanical differences relate to perceived discomfort threshold through preliminary measurements with members of the research team.

\subsection{Design and Setup of a Deformable Pressure Testbed}
\begin{figure*}[t]
    \centering
    \includegraphics[width=\textwidth]{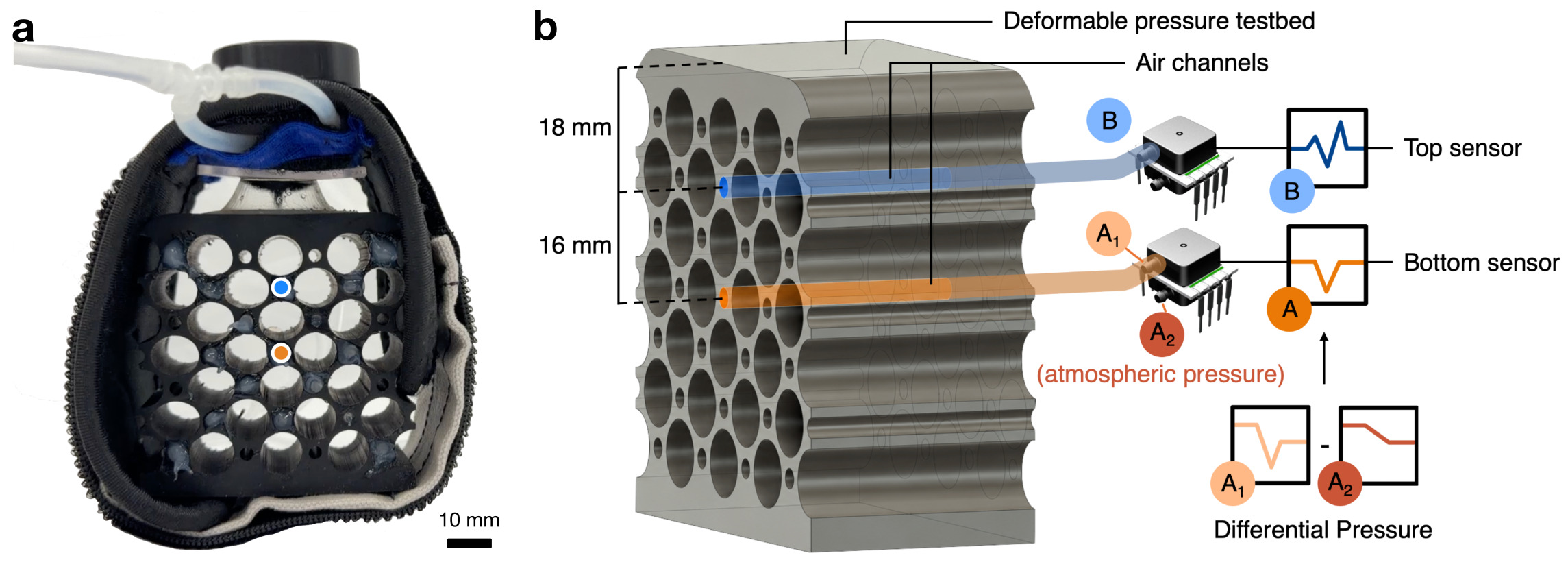}
    \captionsetup{font=footnotesize}
    \caption{\textbf{Deformable pressure testbed setup.}
    (a) Photo of the experimental setup of OUCH wrapped around the deformable pressure testbed, with blue and orange markers indicating top and bottom sensor locations, respectively. 
    (b) Schematic of the deformable pressure testbed, with blue and orange indicating sensor channels for the top and bottom sensors, respectively, and the differential pressure-sensing mechanism used to quantify local pressure changes.}
    \label{fig:set_up}
    \vspace{-6mm}
\end{figure*}

To compare the two indenters on a soft, tissue-like surface, we used a deformable pressure testbed with embedded pressure sensing (Fig.~\ref{fig:set_up}). We 3D printed the testbed from 100\% silicone resin (Silicone~40A, Formlabs) with a Shore hardness of 40A, an ultimate tensile strength of 5~MPa, and an elongation at break of 230\%. We designed the testbed geometry using the metamaterial described in \cite{florijn2014programmable}, a honeycomb-like metamaterial with seven rows of circular channels of two alternating diameters, $D_1 = 3$~mm and $D_2 = 10$~mm, at a pitch of $p = 8$~mm, with overall dimensions of $35 \times 64 \times 68$~mm (width $\times$ length $\times$ height). Under compression, the testbed buckles such that neighboring cells collapse in alternating orthogonal directions and exert compressive forces on one another. %

The internal pressure changes within the testbed were measured using fluidic innervation~\cite{truby2022fluidic}, in which empty air channels are embedded within a 3D-printed structure, and the change in internal air pressure caused by deformation is measured with off-the-shelf pressure transducers. This technique has been used in both wearable applications and sensorized testbeds for clinical applications \cite{FIBiorob,chongIROS}. %
For this study, we sensorized two channels on the vertical centerline of the testbed: a top channel located 18~mm below the contact surface and a bottom channel located 16~mm below the top channel (Fig.~\ref{fig:set_up}(b)). Because each channel is a through-hole, we fully sealed one end of the channel with silicone adhesive (Sil-Poxy, Smooth-On) and sealed the opposite end around a silicone tube that connected the channel to a differential pressure transducer (DLH-L10D, Amphenol All Sensors; $\pm$10~inH$_2$O). %
As the testbed deforms, the internal air pressure of each channel changes, and each transducer measures the pressure difference between its channel and atmospheric pressure, where the second port of each transducer was left open to ambient air (Fig.~\ref{fig:set_up}(b)). The 18-bit digital transducer outputs were read by an ESP32 microcontroller through an I$^2$C multiplexer, converted to pressure using the manufacturer's transfer function, and transmitted to a laptop over Bluetooth at 25~Hz per channel.

During testing, the OUCH cuff was wrapped around the testbed. The indenter, which is attached to the pouches on the inner side of the cuff, was positioned at the center of the top surface of the testbed, directly above the two sensing channels (Fig.~\ref{fig:set_up}(a)). The channels on the vertical centerline were chosen for this reason, so that the sensors lie directly along the loading axis of the indenter. A pneumatic pressure controller (Piezense, Haptica) inflated the pouches through a stepwise profile from 120 to 200~kPa in 20~kPa increments, holding each pressure level for 5~s of inflation followed by 5~s of deflation (Fig.~\ref{fig:deformable_testbed}(a)). Pouch expansion buckled the compliant spacer and drove the indenter into the testbed, compressing the metamaterial and changing the pressure in the sensing channels. The same procedure was repeated with the AlgoMed tip in place of the OUCH indenter.

\subsection{Benchmarking Against a Clinical Algometer}
The following results compare the OUCH indenter with the AlgoMed tip on the deformable testbed, in blocked force, and in discomfort thresholds, supplemented by FEA simulation of OUCH.

\subsubsection{Deformable Testbed Results}
\begin{figure}[t]
    \centering
    \includegraphics[width=0.9\linewidth]{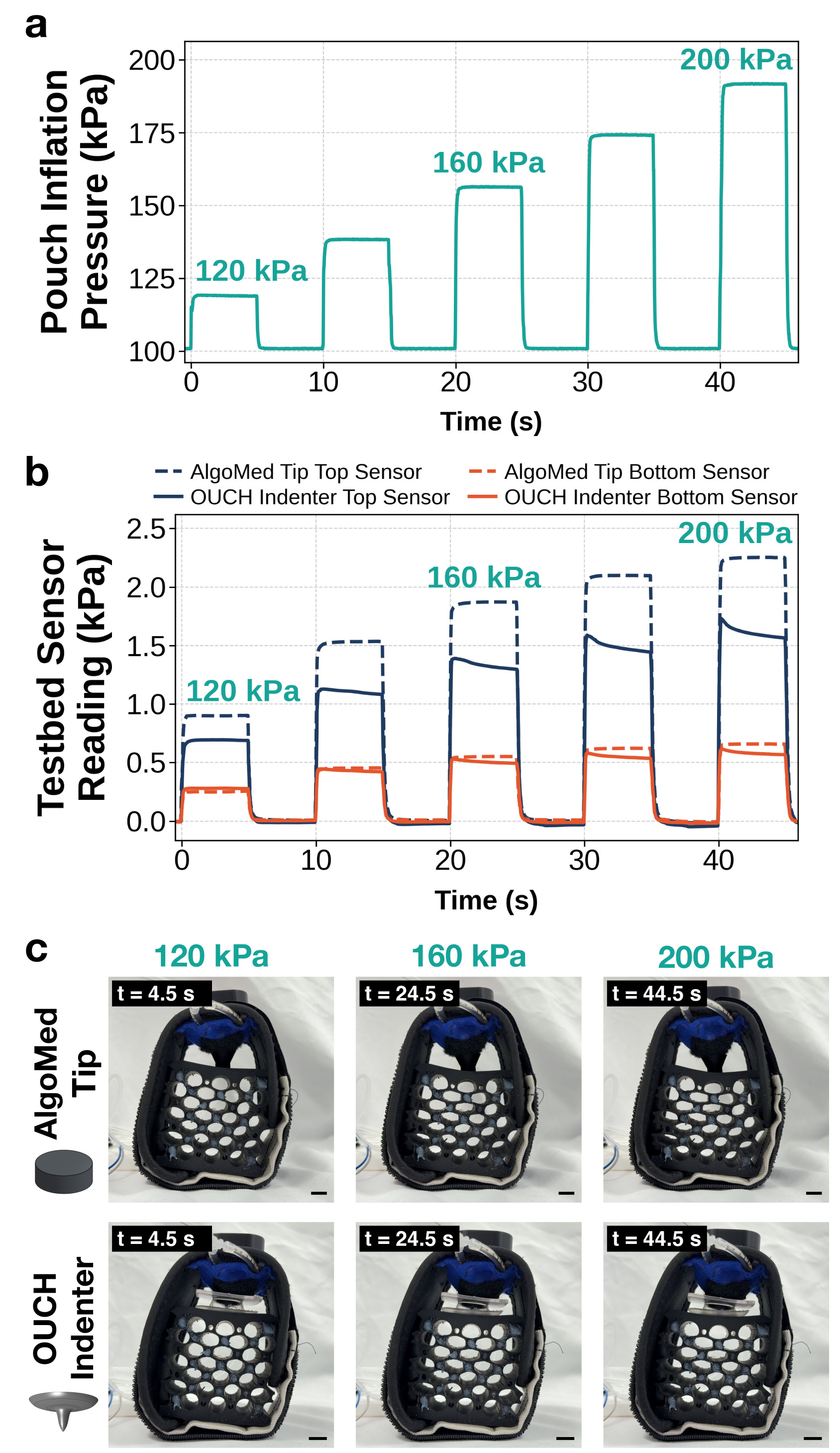}
    \captionsetup{font=footnotesize}
    \caption{\textbf{Deformable pressure testbed characterization.}
    (a) Pouch inflation pressure profile over time during stepwise loading.
    (b) Testbed sensor readings from the top and bottom sensing channels on the deformable testbed indicated in Fig.~\ref{fig:set_up} for the AlgoMed and OUCH indenters under the inflation pressure profile in (a).
    (c) Representative photos of testbed deformation with the AlgoMed and OUCH indenters under pouch inflation pressures of 120, 160, and 200~kPa. Scale bars represent 10 mm.}
    \label{fig:deformable_testbed}
    \vspace{-8mm}
\end{figure}

We first examined how each indenter deformed the testbed near the surface and at depth, as captured by the top and bottom sensing channels embedded within the testbed (Fig.~\ref{fig:deformable_testbed}). %
For both indenters, the testbed sensor readings increased with inflation pressure, and the top sensor consistently recorded higher readings than the bottom sensor. At 200~kPa, the top sensor reached 2.25~kPa with the AlgoMed tip and 1.72~kPa with the OUCH indenter, compared with 0.65~kPa and 0.61~kPa at the bottom sensor, respectively. Because the top sensor lies closer to the contact surface, it experiences greater deformation than the deeper bottom sensor across both indenters. This behavior is similar to human tissue, where deformation decreases from the skin into the deeper subcutaneous tissue. %

The bottom sensor readings were similar for the two indenters, differing by at most 0.04~kPa. The similarity may be attributed to depth, where farther from the indenter contact surface, the effect of the difference in indenter geometry diminishes, and the reading depends mainly on the overall deformation of the testbed produced by the applied load. As shown in Fig.~\ref{fig:deformable_testbed}(c), the circular channels in the bottom rows of the testbed deformed similarly under both indenters. %

In contrast, the top sensor readings differed substantially between the two indenters. Across all inflation pressures, the AlgoMed tip produced higher top sensor readings than the OUCH indenter, with the difference increasing from 0.20~kPa at 120~kPa to 0.53~kPa at 200~kPa. This difference is likely due to indenter geometry. Because each sensor measures the pressure change caused by compression of its air channel, the reading reflects how much of the channel is deformed. The flat AlgoMed tip distributes the load over a contact area of 78.5~mm$^2$, approximately 35 times larger than that of the pointed OUCH indenter (2.27~mm$^2$), and therefore deforms a wider region of the top sensor's channel. This difference is visible in Fig.~\ref{fig:deformable_testbed}(c), where the circular channels in the top rows of the testbed deformed more substantially under the AlgoMed tip than under the OUCH indenter. With increasing depth, the load from either indenter spreads over a wider region, reducing the effect of indenter geometry and resulting in similar bottom sensor readings. Therefore, near the surface, the AlgoMed tip produces broader and greater overall deformation, whereas the OUCH indenter concentrates deformation more locally.

The two indenters also differed in their top sensor temporal response. The AlgoMed tip produced a steady reading during each 5~s hold, whereas the OUCH indenter peaked at the onset of each hold and then gradually declined, with the top sensor reading decreasing by approximately 1\% at 120~kPa and 9\% at 200~kPa. This decline can be attributed to the compliant silicone spacer, which buckles at the onset of inflation to allow the indenter to contact the testbed, but continues to deform gradually during the 5~s hold due to viscoelastic creep of the silicone. As the spacer deforms, the load initially shared by the spacer and the indenter is progressively transferred to the indenter alone, concentrating the load over a smaller area so that less of the testbed surface is being deformed over the hold duration. Consistent with this, the decline was larger at the top sensor than at the bottom sensor (0.16~kPa versus 0.05~kPa during the hold at 200~kPa). Thus, whereas the AlgoMed tip applies a constant load during each hold, the OUCH stimulus becomes increasingly localized as the spacer creeps.

Overall, the two indenters deformed the deeper testbed similarly but differed near the surface, where the OUCH indenter produced a more localized, time-dependent stimulus during each inflation period.

\subsubsection{FEA Simulation on a Deformable Surface}
To examine the contact mechanics that could not be directly observed in the physical deformable testbed, we simulated in FEA how spacer buckling and indenter contact influenced indentation depth, contact area, and contact pressure with increasing compression. The model represented the deformable surface as a nearly incompressible hyperelastic material with a Poisson’s ratio of 0.495 and a fixed lower boundary. It represented the compliant spacer as an isotropic elastic material with a Young’s modulus of 3.5~MPa and a Poisson’s ratio of 0.48, while the rigid indenter was assigned a Young’s modulus of 3500~MPa and a Poisson’s ratio of 0.36. The displacement-controlled analysis applied a 10~mm downward displacement to the assembly, with finite-sliding contact and a friction coefficient of 0.8 defined between interacting surfaces.

\begin{figure}[t]
    \centering
    \includegraphics[width=\columnwidth]{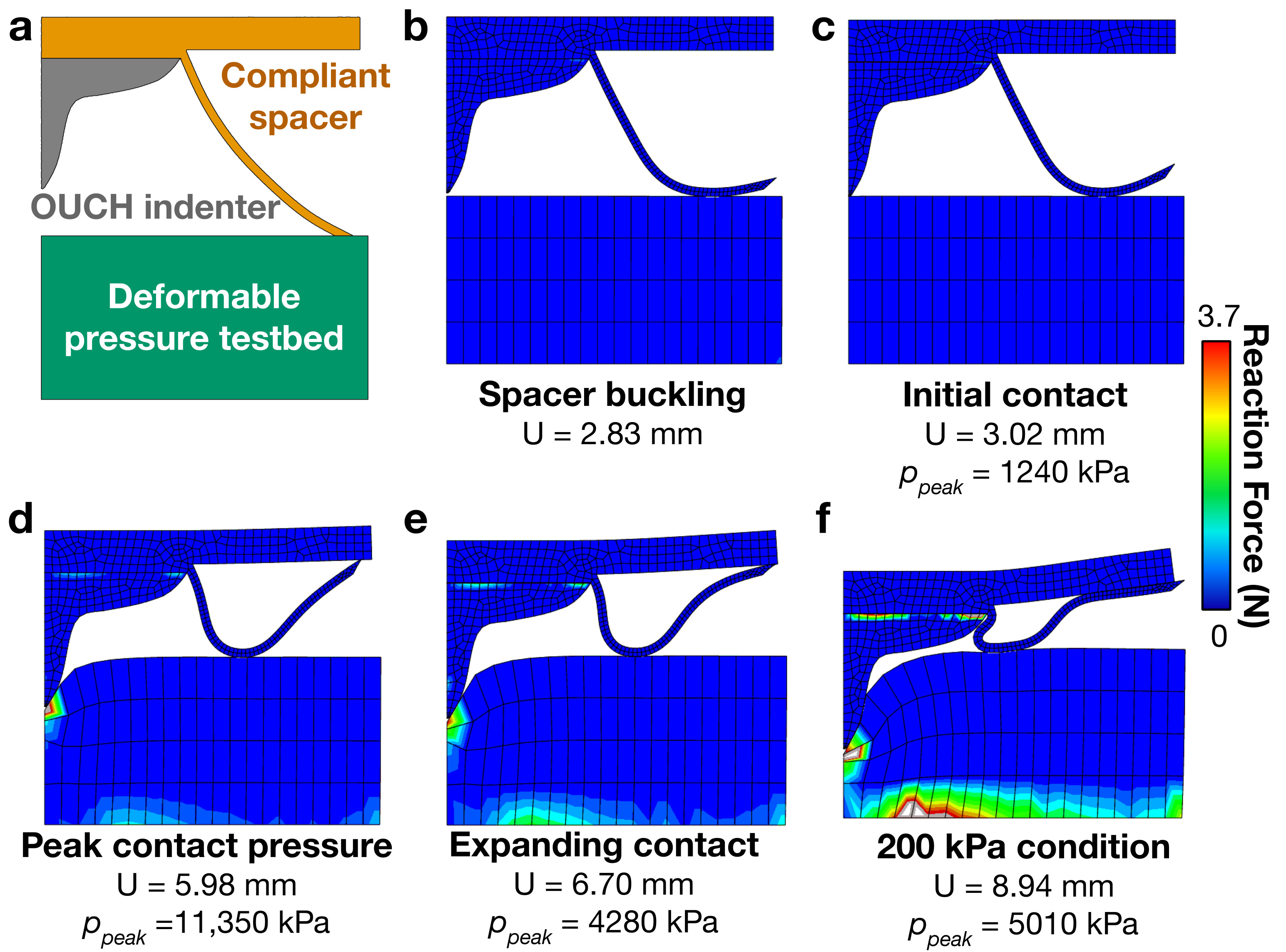}
    \captionsetup{font=footnotesize}
    \caption{\textbf{FEA characterization on the deformable testbed.}
   Cross-sectional FEA results showing spacer deformation and indenter engagement with increasing applied displacement $U$.  
   (a) FEA model of the compliant spacer, OUCH indenter, and deformable pressure testbed.
   (b) Spacer buckling before contact ($U=2.83$~mm).
   (c) Initial contact ($U=3.02$~mm; peak contact pressure 1,240~kPa).
   (d) Maximum peak contact pressure ($U=5.98$~mm; 11,350~kPa).
   (e) Expanded contact ($U=6.70$~mm; 4,280~kPa).
   (f) Loading state corresponding to the 200~kPa testbed condition ($U=8.94$~mm; indentation 5.94~mm; 5,010~kPa).}
    \label{fig:fea_deformable}
    \vspace{-7mm}
\end{figure}

The model provided a cross-sectional view of the interaction among the compliant spacer, the OUCH indenter, and the deformable surface, capturing the transition from spacer deformation and buckling to indenter engagement and testbed indentation (Fig.~\ref{fig:fea_deformable}). As indicated by the local peak in the force-displacement response, spacer buckling occurred at an applied displacement of 2.83~mm, followed by initial indenter contact at 3.02~mm. At first contact, the indenter-testbed contact area was 0.004~mm$^2$, indicating highly localized contact. As displacement increased to 5.98~mm, the contact area increased to 6.96~mm$^2$, while the peak contact pressure reached 11,350~kPa. With further compression, the deformable surface conformed around a larger portion of the indenter, substantially increasing the contact area. At the loading state representing the 200~kPa condition in the physical testbed (applied displacement $U = 8.94$~mm), the indenter penetrated 5.94~mm into the deformable surface, with an indenter-testbed contact area of 150.6~mm$^2$, and a peak contact pressure of 5,010~kPa. This pressure, under deformable surface conditions, was lower than the 18,500~kPa estimated from the blocked force test under the corresponding rigid-surface condition. The difference is consistent with deformation of the testbed and its conformation around the indenter, which increased the effective contact area and redistributed the applied load. This broader load distribution also helps explain why the peak contact pressure decreased after its maximum even as the overall force and indentation continued to increase.

We qualitatively compared video playbacks of FEA simulation and the deformable-testbed experiments to assess whether the model reproduced the observed mechanical behavior of OUCH. Both simulated and real conditions showed progressive spacer deformation followed by indenter contact and increasing testbed deformation under greater loading. Because the spacer was modeled as purely elastic and the analysis was displacement-controlled rather than subjected to sustained loading, the model does not capture time-dependent creep. However, FEA revealed the evolution of contact geometry and pressure distribution on the deformable surface that could not be directly measured on the testbed. It shows that the spacer buckles before the indenter engages the surface, and that, on a compliant surface, the contact pressure produced during stimulation depends on how the surface conforms around the indenter rather than solely on the indenter’s tip area.

\subsubsection{Blocked Force Test Results}
\begin{figure}[t]
    \centering
    \includegraphics[width=0.9\linewidth]{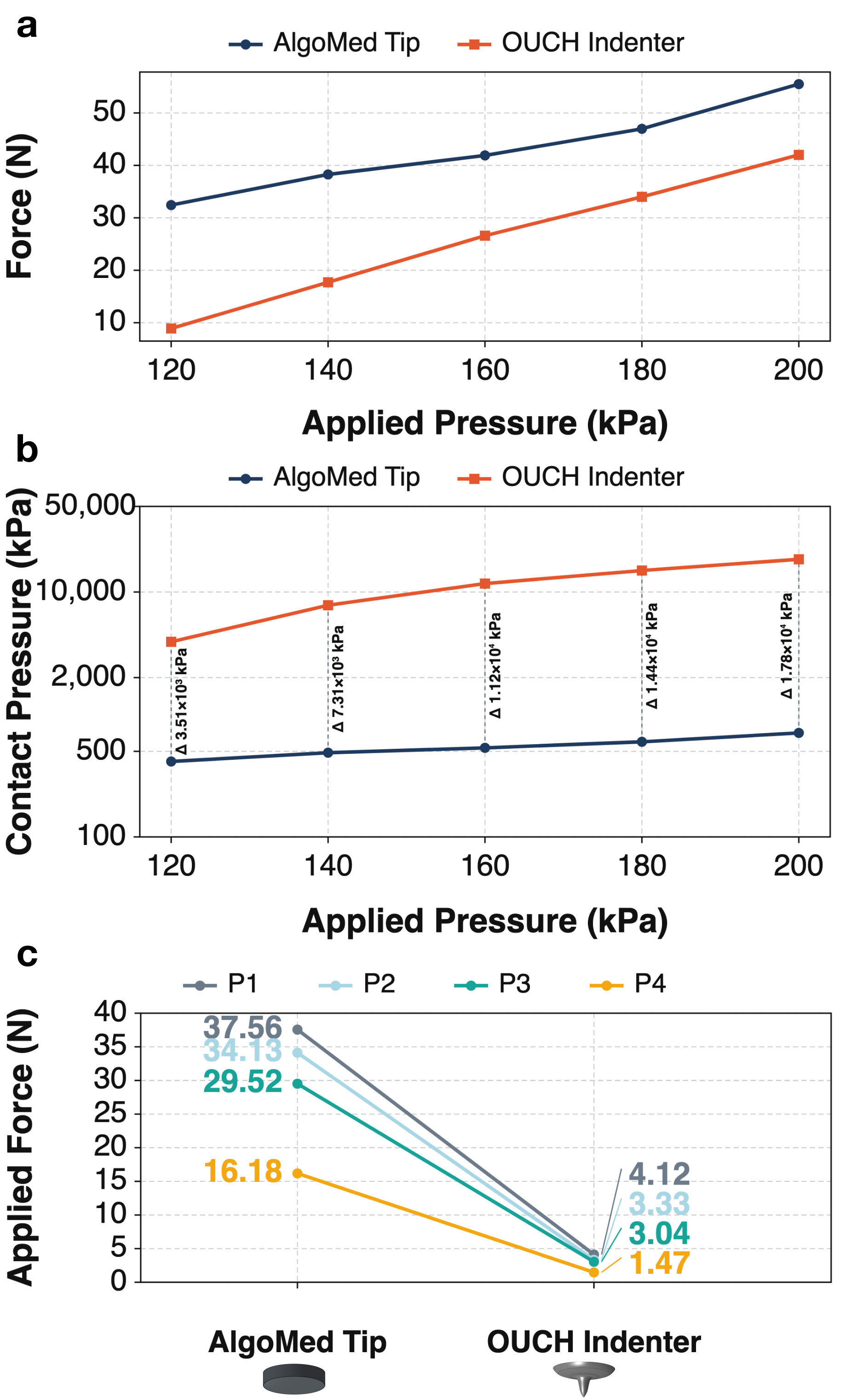}
    \captionsetup{font=footnotesize}
    \caption{\textbf{Comparison of the AlgoMed tip and OUCH indenter.}
    (a) Blocked force measurements comparing force output as a function of applied pressure. 
    (b) Calculated contact pressure on a logarithmic scale using the measured forces in (a) and the respective contact areas of 78.5 mm$^2$ for the AlgoMed tip and 2.27 mm$^2$ for the OUCH indenter, with labels indicating pressure differences between the two tips.
    (c) Applied force at the reported discomfort threshold from four research team members using the AlgoMed tip and OUCH indenter.}
    \label{fig:algomed_comparison}
    \vspace{-8mm}
\end{figure}

To compare the two contact geometries under identical rigid actuation conditions, we attached the AlgoMed tip to the finalized stacked-pouch actuator in place of the OUCH indenter and tested it in the blocked force setup described in Section~\ref{sec:pouch} (Fig.~\ref{fig:pouch_indenter}(a)). As shown in Fig.~\ref{fig:algomed_comparison}(a), the blocked force increased with applied pressure for both the AlgoMed tip and the OUCH indenter, with the AlgoMed tip generally transmitting greater force for the same inflation pressure. At 200~kPa, for example, the AlgoMed tip generated 55.5~N compared with 42.0~N for the OUCH indenter. However, force alone does not capture the contact pressure, which also depends on the contact area. The OUCH indenter has a substantially smaller contact area (2.27~mm$^2$) than the AlgoMed tip (78.5~mm$^2$). At 200~kPa, these forces correspond to estimated contact pressures of 18,500~kPa for the OUCH indenter and 707~kPa for the AlgoMed tip, as shown on the logarithmic scale in Fig.~\ref{fig:algomed_comparison}(b). The OUCH indenter consistently produced substantially higher estimated contact pressure than the AlgoMed tip. The difference increased further with higher applied pressure, reaching 17,800~kPa at 200~kPa inflation pressure. These results demonstrate that the OUCH indenter concentrates the actuator load over a substantially smaller area, producing a more localized contact pressure distribution while reaching a lower overall force.

\subsubsection{Perceived Discomfort Threshold Evaluation}
Finally, to relate these mechanical differences to perception, we conducted a preliminary internal evaluation with four members of the research team to compare discomfort thresholds between the AlgoMed tip and the OUCH indenter. We defined the perceived discomfort threshold as the point at which the sensation first became painful to the participant. For each indenter, we manually increased the applied pressure until the participant verbally indicated ``stop" and recorded the corresponding applied force as the discomfort threshold. As shown in Fig.~\ref{fig:algomed_comparison}(c), the OUCH indenter consistently elicited discomfort at lower applied forces than the AlgoMed tip across all four individuals, with thresholds ranging from 16.18 to 37.56~N with the AlgoMed tip compared with 1.47 to 4.12~N with the OUCH indenter, an approximately 89\% reduction in applied force.

\section{Discussion}
\addtolength{\textheight}{-1cm}
These experiments connect the inflation pressure applied to the actuator to the stimulus delivered at the contact and to the resulting perception. Under blocked force conditions, both indenters produced forces that increased with inflation pressure (Fig.~\ref{fig:algomed_comparison}(a)), which translated to a substantially higher estimated contact pressure of 18,500~kPa for the OUCH indenter at 200~kPa than the AlgoMed tip (707~kPa) (Fig.~\ref{fig:algomed_comparison}(b)). On the deformable testbed, the OUCH indenter produced a top sensor reading 0.53~kPa lower than the AlgoMed tip at 200~kPa, indicating that the OUCH indenter deformed a smaller region near the surface, while the bottom sensor readings differed by at most 0.04~kPa, indicating comparable deformation at depth (Fig.~\ref{fig:deformable_testbed}). Examining this localized contact more closely, the FEA model predicted that the OUCH indenter penetrated 5.94~mm into the deformable surface at 200~kPa, confirming that the localized load produces substantial indentation on a compliant surface. In preliminary perceptual testing, discomfort occurred at 1.47--4.12~N with the OUCH indenter, corresponding to nominal contact pressures of 650--1,810~kPa, compared to nearly an order of magnitude greater threshold of 16.18--37.56~N and substantially lower contact pressures of 210--480~kPa with the AlgoMed tip. This indicates the OUCH indenter relies on higher contact pressure rather than greater required force. When these forces are mapped onto the pressure-force relationships in Fig.~\ref{fig:algomed_comparison}(a), the OUCH indenter required less than the lowest tested inflation pressure of 120~kPa to reach the discomfort threshold, whereas the AlgoMed tip required up to  138~kPa. Although we did not measure nociceptor activation, one interpretation could be that the localized deformation of the pointed OUCH indenter stimulated a small number of nociceptors more quickly or at a higher local amplitude than the flat AlgoMed tip, which spreads a lower-amplitude deformation across a wider region of the surface. This localization offers potential practical advantages for wearable noxious stimulation: the OUCH indenter can evoke discomfort at a lower applied force from the actuator with less overall tissue deformation.

\section{Conclusions and Future Work}
This work presented OUCH, a wrist-worn pneumatic device that delivers controlled noxious mechanical stimulation for neurological assessment. Blocked force testing showed that the stacked-pouch actuator produced predictable force with increasing inflation pressure, and that adding the rigid indenter not only increased the blocked force but also concentrated it over a much smaller contact area. The compliant spacer kept the indenter separated from the skin at rest and allowed contact under actuation. Benchmarking against a clinical pressure algometer showed that, at the same applied pressure, the OUCH indenter produced substantially higher contact pressure. On the deformable pressure testbed, both indenters deformed the deeper testbed layer similarly, but near the testbed surface, the algometer produced greater deformation over a wider region, whereas the OUCH indenter concentrated localized deformation in a smaller area. As a result, in preliminary testing, the OUCH indenter reached the discomfort threshold at a lower applied force. Together, these results suggest that localizing the load, rather than applying greater force, is an effective strategy for evoking discomfort with a compact wearable actuator.

Several limitations remain. Discomfort thresholds were obtained from preliminary testing within the research team, and the viscoelastic creep of the spacer observed during inflation was not characterized and could be addressed in future designs. Additionally, the deformable testbed approximates, but does not replicate, the biomechanics of human skin and subcutaneous tissue. More broadly, characterizing the device required multiple complementary setups, each with its own trade-offs: the custom deformable testbed better approximates soft-tissue contact but requires FEA to estimate indentation and contact pressure, whereas the blocked force test is standard and repeatable but less realistic for on-wrist behavior. Because blocked force is measured with the pouch fully constrained, the force delivered to the wrist from a wearable cuff is likely lower, and the inflation pressure required in practice may be somewhat higher. Nevertheless, our results suggest that the OUCH indenter can evoke discomfort below the low end of the tested pressure range.

Future work includes a formal user study with healthy individuals to verify that OUCH’s operating range produces sufficient discomfort while remaining below reported thresholds for skin injury, then proceeding with clinical study in patients with disorders of consciousness to evaluate evoked motor responses. We are also interested in exploring how perceived discomfort relates to localized skin surface load rather than overall tissue deformation, as well as how different skin conditions influence force-pain thresholds. Our overall goal is to integrate OUCH with accelerometer-based motion tracking to deliver standardized stimuli and objectively quantify the evoked motor response, enabling automated neurological assessment in place of manual, subjective bedside examination in active hospital environments.
\ifblindreview
\else
\section*{Acknowledgment}
We thank Ereni Delis for providing insights on pain thresholds and Nitin Seshadri for providing clinical feedback on the device design.
\fi

\bibliographystyle{IEEEtran} %
\bibliography{references} %
\end{document}